\documentclass[11pt]{article}

\usepackage[final]{acl}

\usepackage{times}
\usepackage{latexsym}
\usepackage{booktabs}
\usepackage[T1]{fontenc}

\usepackage[utf8]{inputenc}
\usepackage{amsmath}
\usepackage{amssymb}

\usepackage{microtype}

\usepackage{inconsolata}

\usepackage{tabularx}
\usepackage{graphicx}
\usepackage{xspace}
\title{\MultiVLM: Prior-Steered Multi-View VLM Alignment for Hallucination-Robust Table OCR}

\author{
 \textbf{Guangyi Liu\textsuperscript{1}},
 \textbf{Qianjun Huang\textsuperscript{1}},
 \textbf{Boyu Hou\textsuperscript{1}},
\\
  \textsuperscript{1} Huawei Technologies, Co., Ltd 
\\
 \small{
   \textbf{Correspondence:} \href{mailto:email@domain}{liuguangyi5@huawei.com; guangyi@udel.edu}
 }
}

\newcommand{\modeltheme}{\textsc{Doc}}

\newcommand{\ModelA}{{\modeltheme}\textsc{Expert-1B}\xspace}

\newcommand{\MultiVLM}{PrismAlign\xspace}

\begin{document}

\maketitle
\begin{abstract}

Table extraction suffers from frequent structural errors and semantic hallucinations. We propose \MultiVLM, a multi‑VLM framework aligning diverse visual perspectives to resolve ambiguity. It integrates priors of table logic to assess output plausibility, decoupling structural alignment from cell content alignment. A Bayesian decision strategy maximizes alignment accuracy by exploiting the correlation between extraction errors and computable rule violations. Evaluated on open-source and custom VLMs, \MultiVLM  reduces hallucinations and achieves state-of-the-art performance on OmniDocBench 1.5, as well as on the table category of CC‑OCR and PureDocBench.

\end{abstract}

\section{Introduction}

As Large Language Models (LLMs) penetrate diverse industries, converting unstructured documents---reports, forms, contracts, manuals---into language-grounded, LLM-consumable knowledge, has become critical \cite{liu2024focus, wei2025deepseek, niu2025mineru25decoupledvisionlanguagemodel, olmocr2}. This conversion directly supports two key objectives: generating high-quality domain-specific data for LLM training and fine-tuning \cite{deepseekai2025deepseekr1incentivizingreasoningcapability, Qwen3}, and constructing knowledge bases that empower LLM-driven agents to retrieve, reason, and act upon real-world content \cite{Zep, ADK}.

Recent advances in Vision-Language Models (VLMs) have fundamentally reshaped document understanding \cite{2302.05442, dosovitskiy2020image, wei2024got, cui2025paddleocr30technicalreport, feng2025dolphin, 2307.06304, zhu2025internvl3, hunyuanvisionteam2025hunyuanocrtechnicalreport, wang2025infinityparserlayoutaware}. 
Autoregressive decoding, in particular, has proven effective for handling variable-length outputs such as  formulas in LaTeX and  tables in HyperText Markup Language (HTML). 

However, the extraction frequently suffers from hallucinations, generating non-existent cells, misinterpreting table hierarchies, or fabricating symbols that are not present in the source image. A significant barrier to progress lies in the fragmentation of training resources; while leading models utilize their own proprietary data, the lack of unified open-source datasets means that each model develops distinct biases and strengths. 

Recent work has explored multi‑model frameworks for improving text recognition through ensemble agreement verification to mitigate individual weaknesses \cite{2026CVPR}. However, applying the method directly to table extraction is suboptimal. Unlike free-form text, tables possess inherent structural rigidity and logical constraints.

To address this, we propose \MultiVLM, a  multi-view VLM  framework inspired by stereopsis to align disparate visual perspectives to resolve ambiguity. 
The framework leverages  prior knowledge of table structures and semantics to evaluate the plausibility of each VLM's output. Since structural errors and cell content errors exhibit distinct patterns, 
table structural alignment and cell content alignment are decoupled.
Utilizing the correlation between VLM extraction errors and some computable rule violation scores, a Bayesian decision approach \cite{Bayesiancombo} is applied to maximize the posterior alignment accuracy.

With multiple VLMs input to \MultiVLM, including a custom-built 1B model optimized for efficient inference,  non-trivial improvements is achieved in table extraction hallucination robustness.
Notably, our approach attains state-of-the-art (SOTA) performance on  OmniDocBench 1.5 \cite{ouyang2024omnidocbenchbenchmarkingdiversepdf}, as well as on the table subsets of CC-OCR \cite{ccbench}, and PureDocBench \cite{li2026puredocbench}.
The contributions are threefold:
\begin{itemize}
    \item We introduce \MultiVLM, a multi-view vision–language framework that aligns heterogeneous VLM outputs by explicitly modeling structural and semantic consistencies across views. The framework resolves ambiguous table regions with cross-model consensus rather than relying on any single model’s prediction.

    \item We decouple table parsing into structural alignment and cell-content alignment, and formulate table selection as a Bayesian decision problem that maximizes the posterior likelihood of correctness.

    \item We demonstrate that \MultiVLM significantly improves robustness against table-related hallucinations. \MultiVLM achieves new SOTA results on OmniDocBench 1.5, as well as on the table subsets of CC-OCR and PureDocBench. A custom-built VLM, \ModelA, is developed and applied in \MultiVLM to improve inference efficiency.
\end{itemize}

\section{Related Work}
\label{sec:related} 

\begin{sloppypar}
The field of document intelligence has undergone a fundamental transformation, shifting from sequential, modular Optical Character Recognition (OCR) pipelines to document parsing driven by specialized VLMs \cite{wei2024got, wei2025deepseek, cui2025paddleocr30technicalreport}. This evolution is characterized by the need to comprehend not just text, but the complex visual and logical structure of  documents. Moving beyond the semantic fluency often exhibited by general-purpose VLMs \cite{Qwen3}, the evolution requires deep understanding of visual grammar that governs element positioning, spatial hierarchy, and reading order, as well as a coherent representational framework for text, tables, formulas, and charts.

\subsection{Design Tradeoffs}

This subsection surveys popular architectural paradigms: pipeline-based multi-model systems,  end-to-end unified models, and two-stage hybrid approaches. Each presents distinct tradeoffs in accuracy, efficiency, scalability, and error dynamics.

Pipeline-based frameworks decompose document parsing into discrete stages---layout analysis, text recognition, table and formula parsing---each handled by specialized models. This modularity allows practitioners to reuse mature, domain-optimized tools, reducing development overhead and benefiting from decades of OCR improvements. However, this design inherently suffers from error accumulation. Mistakes in early stages, such as incorrect layout segmentation, propagate downstream, often leading to compounding inaccuracies. Additionally, these systems require extensive postprocessing logic to reconcile outputs across modules, increasing engineering complexity. Nonetheless, certain frameworks deliberately embrace this paradigm to maximize reliability in high-stakes domains. For example, DianJin-OCR-R1 \cite{dianjin-ocr-r1} adopts a  strategy where a primary VLM invokes specialized expert models as external tools. 

In contrast, unified architectures process documents within a single, monolithic VLM, eliminating modular boundaries. These end-to-end models offer significant advantages in system simplicity, ease of deployment, and avoidance of cascading errors. By removing intermediate stages, they reduce operational overhead and streamline postprocessing. However, this elegance comes with challenges. Dense-text documents generate long token sequences, placing pressure on attention mechanisms and increasing the risk of information loss or hallucination. Models like DeepSeek-OCR \cite{wei2025deepseek} and GOT-OCR \cite{wei2024got} 
demonstrate that architectural unification can rival or exceed pipeline-based frameworks.

Two-stage architectures occupy a middle ground, explicitly decoupling layout detection from content extraction. This paradigm can manifest in two forms: single-model two-stage systems, such as MinerU 2.5 \cite{niu2025mineru25decoupledvisionlanguagemodel}, which internally separate layout refinement from content generation via anchor prompts; and multi-model two-stage systems, such as PaddleOCR-VL 1.5 \cite{cui2025paddleocr30technicalreport}, which employ distinct models for each stage. These approaches reduce perceptual ambiguity before textual decoding, improving structural fidelity. However, they still require moderate postprocessing and remain susceptible to  error propagation.

\subsection{Core Architectural Strategies for Efficiency and Fidelity}

Architecturally, these systems adopt advanced  vision encoders. Earlier works like Dolphin \cite{feng2025dolphin} employ Swin Transformer, while later systems such as MinerU 2.5 and PaddleOCR 3.0 integrate NaViT \cite{2307.06304} to handle dynamic image resolutions and varying aspect ratios. To further improve efficiency, DeepSeek-OCR optimizes tile-based vision encoders  \cite{zhu2025internvl3} using pretrained SAM and CLIP models, significantly reducing vision token counts. DeepSeek-OCR 2 \cite{wei2026deepseek} introduces learnable casual query module in the vision encoder to dynamically reorder visual tokens upon image semantics.

HunyuanOCR \cite{hunyuanvisionteam2025hunyuanocrtechnicalreport} introduces XD-Rotary Positional Embedding (RoPE), which deconstructs the conventional RoPE into four independent subspaces: text, height, width, and time. This establishes a native alignment mechanism bridging 1D text sequences, 2D page layouts, and 3D spatiotemporal information, enhancing structural understanding.

 \end{sloppypar}

\section{Method}

A table dataset is denoted as: $\mathcal{E} = \{\mathcal{E}_1, \dots, \mathcal{E}_N\}$, where each  $\mathcal{E}_i$  is a table element. In this section, we focus on table, but the approach also applies to formula. $\mathcal{V} = \{V_1, \dots, V_K\}$ denotes the set of VLMs for table extraction, and for $\mathcal{E}_i$, each VLM $V_k$  produces a candidate extraction: $\hat{Y}_{k,i}$.

\begin{figure*}[htbp]
    \centering
    \vspace{-10pt}
\includegraphics[width=1\textwidth]{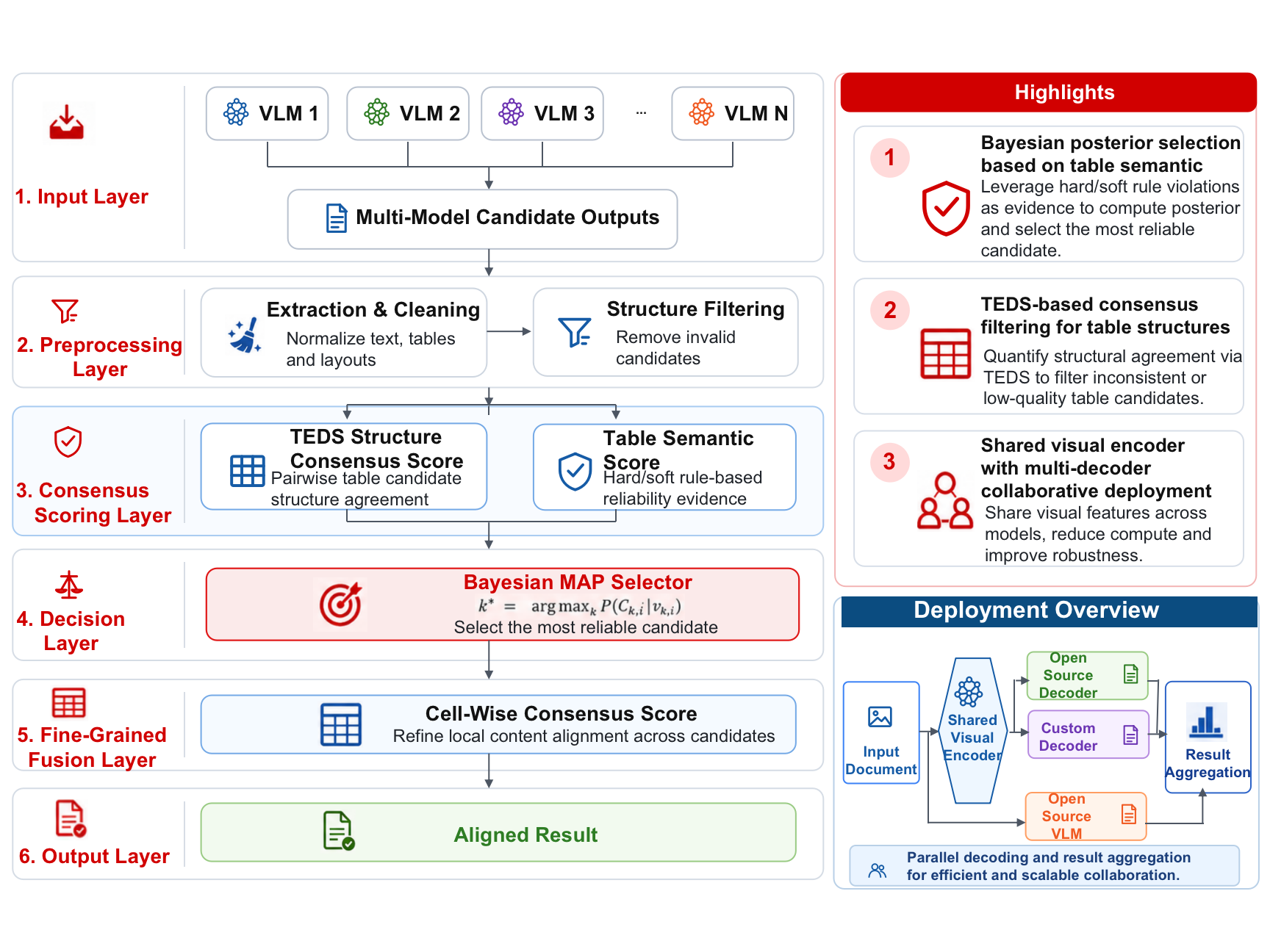} 
\vspace{-40pt}
\caption{Architecture of \MultiVLM}
\label{fig:architecture}
\end{figure*}

A six-layer architecture is adopted comprising of input layer, preprocessing layer, scoring layer, decision layer, fine-grained fusion layer, and output layer, as shown in Fig. \ref{fig:architecture}.
In the input layer, $K$ VLM outputs are collected, each representing a candidate table normalized into HTML format from various source format.

The preprocessing layer performs two main operations, text cleaning and table structure filtering. Cleaning table contents includes removal of redundant whitespace, normalization of symbols, e.g., parentheses, ticks, crosses, and standardization of formula encoding. Table structure filtering means elimination of structurally invalid candidates according to predefined schema constraints. Structured formats such as HTML provide explicit syntactic rules that can be validated. Also, tables embody implicit structural invariants that extend beyond surface markup; for instance, each row must preserve a consistent number of columns, maintaining matrix integrity (see Table \ref{strucuture} in the Appendix for more examples).

\setlength{\floatsep}{4pt}
\setlength{\textfloatsep}{4pt}
\setlength{\intextsep}{4pt}

\subsection{Table Semantic Scoring}

Semantic coherence shapes what constitutes a plausible table. Meaningful tables rarely exhibit degenerate artifacts, such as entirely empty header rows, isolated numeric values without contextual labels, or structurally valid but semantically void “orphan digits.” In a well-formed table, every numerical entry is semantically anchored to both its row and column headers; without these descriptors, the value becomes informationally meaningless.

Additionally, 
column homogeneity requires that cells within a column share consistent semantic types (e.g., numeric, temporal, categorical), preventing intermingling of unrelated data classes that signal boundary misalignment. Similarly, tables exhibiting excessive empty cells relative to their total volume sometimes reflect misaligned parsing. Extraordinary row or column counts (e.g., hundreds of rows from a single document page) typically indicate duplicated or fragmented extraction.

Let $G^S$ denote a set of table semantic rules. 
Given VLM $k$ and table $i$, for rule $g_m$, $0<m\le M$, where $M = |G^S|
$, define a binary  indicator:
\begin{align*}
v_{k,i,m} =
\begin{cases}
1 & \text{if rule } g_m \text{ is violated}, \\
0 & \text{otherwise}.
\end{cases}
\end{align*}
Crucially, all rule evaluations are programmatically computable, providing deterministic, interpretable input for downstream Bayesian inference. More examples can be referred in Table \ref{semantics} in the appendix.

\subsection{TEDS-Structure  Consensus Scoring}

Document parsing differs from generic visual question answering (VQA) or image captioning in a fundamental respect: OCR outputs are inherently deterministic \cite{youtu-parsing}. Under a normalized representation, each input image maps to a unique ground-truth structure---whether plain text, character-level detection and matching (CDM)-rendered formulas, or HTML-formatted tables. Empirical studies indicate that correct OCR results produced by different models tend to converge. Incorrect results exhibit model-specific divergence; frequently, the error patterns are different across models \cite{2011ICDAR, 2026CVPR}. This determinism introduces a new design dimension: agreement across independently generated VLM outputs implies high confidence in correctness, while prediction divergence serves as a strong indicator of potential errors.

Denoting $g_0$ as the consensus rule: the majority of the TEDS-Structure \cite{ouyang2024omnidocbenchbenchmarkingdiversepdf} differences between $~\hat{Y}_{k,i}$ and the set $\{\hat{Y}_{j,i}\}$, for $1\le j \le K$ and $j\neq k$, should be bigger than a threshold $\epsilon$.
For a candidate extraction $\hat{Y}_{k,i}$, we define a score indicator

\begin{align*}
v_{k,i,0} =
\begin{cases}
\;1 & \text{If rule}~ g_0~ \text{is violated}. \\[1.2ex]
\;0 & \text{otherwise}.
\end{cases}
\end{align*}

For $\hat{Y}_{k,i}$, the complete score profile is:
\[
\mathbf{v}_{k,i} = \bigl[v_{k,i,0}, v_{k,i,1},\; v_{k,i,2},\; \dots,\; v_{k,i,M}\bigr]
\]

\begin{table*}[htbp]
    \centering
    \small
    \begin{tabularx}{1\textwidth}{lcccccc}
            \toprule
            \textbf{VLM} & \textbf{Size} & \textbf{Overall↑} & \textbf{TextEdit↓} & \textbf{FormulaCDM↑} & \textbf{TableTEDS↑} & \textbf{TableTEDS-S↑} 
            \\
            \midrule
            Nanonet 2 \cite{Nanonets-OCR-S} & 3B &	83.13 &	0.114 &	78.02 &	82.77 &	87.75 
            \\
            MinerU 2.5 &	1.2B	& 90.67 &	0.047 &	88.46 &	88.22 &	92.38 
            \\
      
            dots.ocr \cite{li2025dotsocrmultilingualdocumentlayout}	& 3B	& 88.41 &	0.048 &	83.22 &	86.78 &	90.62 
            \\
            MonkeyOCR-3B \cite{monekyocr}	& 3B &	87.13 &	0.075 &	87.45 &	81.39 &	85.92 
            \\ 
            Deepseek-OCR	& 3B	& 87.01 &	0.073 &	83.37 &	84.97 &	88.80 \\
            
            Youtu-Parsing & 2.4B & 93.22 & 0.045 & 93.19 & 91.15 & 94.53 \\
            MinerU2-VLM &	0.9B &	85.56 &	0.078 &	80.95 &	83.54 &	87.66 
            \\
            olmOCR \cite{olmocr2} &	7B &	81.79 &	0.096 &	86.04 &	68.92 &	74.77 
            \\

            Dolphin-v2 \cite{feng2025dolphin} & 3B &	89.78 	& 0.054 &	87.63 &	87.02 &	90.48 
            \\            
            POINTS-Reader \cite{points-reader} &	3B &	80.98 &	0.134 &	79.20 &	77.13 &	81.66 
            \\
            PaddleOCR-VL 1.5  &	0.9B &	94.50 &	0.035 &	94.21 &	92.76 &	95.79 
            \\
            
            FD-RL & 4B & 90.41 & 0.049  & 88.67 & 87.35 & 92.10  \\
            
            \midrule
            \ModelA & 1B & 92.26 & 0.046  & 89.09 & 93.375& 96.17  
            \\
                        \MultiVLM & 3.1B & \textbf{95.14} & 0.035  & 94.31 & \textbf{94.62} & \textbf{97.30} 
                        \\            
     
        \bottomrule
        \end{tabularx}
        \vspace{5pt}
        \caption{Comparison on OmniDocBench 1.5}
        \label{tab:OmniDocBench}
    \end{table*}

\subsection{Bayesian Estimate and Approximations}

Given the extraction result of multiple VLM models, the goal is to choose the one with maximum computed posterior probability given $\mathbf{v}_{k,i}$; i.e.,  select $k^*_i$ such that 
   $$
   k^*_i = \arg\max_k P(C_{k,i} \mid \mathbf{v}_{k,i}),$$
where $C_{k,i}$  denotes the event that  $\hat{Y}_{k,i}$ is correct, and  $P(C_{k,i} \mid \mathbf{v}_{k,i})$ is posterior probability that the extraction is correct given $\mathbf{v}_{k,i}$. 

By Bayes’ rule:
$$P(C_{k,i} \mid \mathbf{v}_{k,i}) =
\frac{
P(\mathbf{v}_{k,i} \mid C_{k,i})\; P(C_{k,i})
}{
P(\mathbf{v}_{k,i})
},$$
where $P(\mathbf{v}_{k,i} \mid C_{k,i})$ is the likelihood of observing the score profile $\mathbf{v}_{k,i}$ if the extraction is correct.

To estimate, a validation  dataset is constructed by randomly sampling from publicly available table dataset, denoted as $\Theta_{t}$.
With $\Theta_{t}$, $\pi_m = P(v_{k,i,m} = 1 \mid C_{k,i}), 0 < m \leq M,$ can be estimated, which quantifies how often each table semantic rule is violated, even when the extraction is correct. Note that $\pi_m$ is independent of the VLM being used.

For each VLM $V_k$, we perform inference on the validation datasets $\Theta_t$. 
Then, per-VLM correction rate, $\bar{P}_k$, can be computed by comparing the inference results and the dataset ground truth, and average across $\Theta_t$; and we use $\bar{P}_k$ to estimate $P(C_{k,i})$. 

Also, $\pi_{0, k} = P(v_{k,i,0} = 1 \mid C_{k,i})$ can be computed by comparing the inference outcomes of different VLMs, and average across $\Theta_t$. By designing table semantic rules to minimize mutual correlation,  
we assume rule violations are conditionally independent; then,
\begin{align*}
&P(\mathbf{v}_{k,i} \mid C_{k,i}) =
\prod_{m=0}^M P(v_{k,i,m} \mid C_{k,i}) = \pi_{0, k}^{\,v_{k,i,0}} \\
&(1 - \pi_{0, k})^{1 - v_{k,i,0}} \prod_{m=1}^M \pi_{m}^{\,v_{k,i,m}} (1 - \pi_{m})^{1 - v_{k,i,m}}.
\end{align*}

Concurrently, for the consensus rule and every rule in $G^S$, we estimate its empirical violation rate with VLM $V_k$, \[
p_{k,m} = \dfrac{1}{\lvert\Theta_{t}\rvert}\sum_{i = 1}^{\lvert\Theta_{t}\rvert}\mathrm{id}_{\{v_{k,i,m} = 1\}}, 0 \leq m \leq M,
\]
computed over all elements in the corresponding validation set, where $\lvert \cdot \rvert$ denotes cardinality, and $\mathrm{id}_{\{\cdot\}}$ denotes the indicator function.

When $M$ is small, Monte Carlo simulation can be used to estimate $P(\mathbf{v}_{k,i})$. When $M$ is large, under conditional independence assumption, we estimate \( P(\mathbf{v}_{k,i}) \) to be
   $\bar{P}(\mathbf{v}_{k,i}) =
   \prod_{m=0}^M p_{k,m}^{\,v_{k,i,m}} (1 - p_{k,m})^{1 - v_{k,i,m}}.$
This allows us to compute the exact Bayes denominator per VLM. When the distribution of the benchmark dataset is different from the distribution of the validation dataset, this estimation can be inaccurate. 

When SOTA VLM models are used, $\pi_m$ is close to 0 in practice; i.e. $
\log(1 - \pi_m) \approx -\pi_m
\quad\text{and}\quad
\log \pi_m < 0$. Throughout this paper, log($\cdot$) denotes the natural logarithm.

Define the log-likelihood penalty as
\begin{align*}
    &\ell_{k,i}
= \log P(C_{k,i} \mid \mathbf{v}_{k,i})
=\sum_{m=1}^M
v_{k,i,m}\, \log \pi_m \; \\
&-
\sum_{m=1}^M
(1-v_{k,i,m})\pi_m +
\log \bar{P}(C_{k,i}) - \log \bar{P}(\mathbf{v}_{k,i}) \\
& + v_{k,i,0}\, \log \pi_{0, k} -
(1-v_{k,i,0})\log(1-\pi_{0,k})
\end{align*}
When $\pi_m$ is small, the second term can be neglected, 
and each violated rule contributes a penalty proportional to  $\log \pi_m$.
Therefore, the posterior probability is log-dominated by the VLM candidate with the smallest cumulative rule penalty. Notably, for a rule $g_m$ with prior $\pi_{m} \to 0$, the corresponding penalty term $
\log \pi_{m} \to -\infty$ grow unbounded. This implies that violating a rarely-broken rule incurs a far greater penalty than violating multiple commonly-broken ones. 
\\

When the chosen candidate using the above Bayes approach shares  identical table structure with other candidates, within the fine-grained fusion layer,  pairwise cell-level comparison is performed. Table cells exhibiting inconsistent content across candidates are discarded to improve reliability. Specifically, let $s_1$, $s_2$, $s_3$ be extracted cell-content strings from the three VLMs for the same table cell, and $\mathbb{E}(\cdot, \cdot)$ be normalized edit distance.  Then, $s_1$ is discarded if
$\mathbb{E}(s_1, s_2) + \mathbb{E}(s_1, s_3) < \mathbb{E}(s_1, s_2) + \mathbb{E}(s_2, s_3)$, and $\mathbb{E}(s_1, s_2) + \mathbb{E}(s_1, s_3) < \mathbb{E}(s_1, s_3) + \mathbb{E}(s_2, s_3)$, in which case
the final output is randomly chosen from the set \{$s_2$, $s_3$\}.

Finally, the output layer provides the selected table along with its associated confidence score. In practical deployments, tables with low confidence can be prioritized for manual review, while high-confidence results may be accepted automatically. This confidence score can be computed as the weighted average of $1-v_{k^*_i,i,0}$ and $ P(C_{k^*_i,i} \mid \mathbf{v}_{k^*_i,i})$, where $k^*_i$ is the chosen candidate.

\section{Experiment}

\subsection{Light-Weight VLM}

The architecture of most VLMs are optimized for diverse VQA tasks, which typically demand open-ended outputs. This objective differs fundamentally from OCR, where results impose a deterministic mapping of the input as well as its encoded visual features. This characteristic reduces the necessity for deep decoders, where most inference latency in autoregressive ViT models comes from.

Hence, we adopt the MinerU 2.5 Pro encoder, apply a custom decoder, and introduce VLM \ModelA. \ModelA is trained with 12 M open-source examples, and as shown in Table \ref{tab:OmniDocBench}, achieves competitive performance on OmniDocBench 1.5. The decoder has 16 hidden layers, which is as far as we know the smallest among  VLMs with TEDS score above 90 on OmniDocBench. This significantly reduces  KV-cache requirement, and achieves 1.5× faster inference compared to MinerU 2.5 under identical conditions, making it an efficient component in \MultiVLM\footnote{\ModelA has been made publicly available.  https://huggingface.co/linglongOCR-group/DocExpert-1B.}.

\subsection{Benchmark Result}

In our experiments, \MultiVLM integrates three VLMs in the input layer:  PaddleOCR-VL 1.5 \cite{cui2026paddleocrvl15multitask09bvlm},  MinerU~2.5, and  \ModelA. 
A validation dataset with size $\lvert \Theta_t \rvert = 10,0000$, sampled from PubTable-1M dataset \cite{smock2022pubtables}, is used to estimate the Bayes factors to compute the posterior probability.
We evaluate the proposed approach on three benchmarks: OmniDocBench~1.5, CC-OCR, and PureDocBench. As shown in Tables~\ref{tab:OmniDocBench}, \ref{CC-OCR}, and \ref{puredocbench}, \MultiVLM consistently outperforms all individual component VLMs in terms of table TEDS. Compared with MinerU~2.5, \MultiVLM improves TEDS by 6.40, 14.91, and 9.23 points across the three benchmarks; relative to PaddleOCR-VL 1.5, the improvements are 1.86, 1.36, and  7.23 points, respectively.

Notably, \MultiVLM achieves SOTA performance on table parsing, surpassing FD-RL \cite{zhong2025reading} on PureDocBench and Youtu-Parsing \cite{youtu-parsing} on CC-OCR by substantial margins. When extending heuristic multi-candidate filtering strategy to mathematical formulas with latex keyword, \MultiVLM also attains the best overall performance on OmniDocBench 1.5. These results demonstrate that multi-model selection based on prior knowledge of table structures and semantics can be an effective approach to manage VLM hallucination and to improve extraction accuracy.
\\

To isolate the contribution of individual VLM model, we conducted an ablation study by disabling DocExpert-1B, which simultaneously deactivates  table structure consensus score and cell-wise consensus score. Under this setting, PrismAlign still achieves gains of 1.182 on TableTEDS; this indicates that table structure filtering and table semantic scoring constitute the primary sources of improvement, while other factors further contribute additional positive gains. 

As shown in the deployment overview in Fig. \ref{fig:architecture}, when applying PaddleOCR-VL 1.5, \ModelA and MinerU 2.5 Pro simultaneously to \MultiVLM, the latter two VLMs share the same encoder, and then only one copy of the encoder is deployed during inference. From experiment, slightly sacrificing OmniDocBench accuray, replacing MinerU 2.5 with MinerU 2.5 Pro reduces the total deployed parameters for \MultiVLM from 3.1B to 2.425B.
\\

 \begin{table}[t]
    \centering
\label{tab:html_latex_rules}
    \begin{tabular}{l | p{0.4\linewidth}}
    \toprule
    \textbf{VLM} & \textbf{TableTEDS↑} \\
    \midrule
    MinerU 2.5 & 67.96\\
    PaddleOCR-VL 1.5 & 81.51\\
    Youtu-Parsing & 81.37 \\
    \MultiVLM & \textbf{82.87}\\
    \bottomrule
    \end{tabular}
    \caption{CC-OCR table result}
        \label{CC-OCR}
    \end{table}

   \begin{table}[t]
    \centering
    \normalsize
    \vspace{12pt}

    \begin{tabular}{l | p{0.4\linewidth}}
    \toprule
    \textbf{VLM} & \textbf{TableTEDS↑} \\
    \midrule
    MinerU 2.5 & 73.6 \\
    PaddleOCR-VL 1.5 & 75.6\\
    FD-RL & 80.50 \\
    \MultiVLM & \textbf{82.83}\\
    \bottomrule
    \end{tabular}
    \caption{PureDocBench table result}
\vspace{12pt}\label{puredocbench}
    \end{table}

\section{Conclusion}

In this work, we address the persistent challenge of table extraction hallucinations in vision‑language models by introducing \MultiVLM, a multi‑view framework that aligns diverse visual perspectives. Departing from generic multi‑model ensembles, our approach explicitly accounts for the structure and semantics constraints of tables, decoupling structural consensus from cell content consensus, and leveraging Bayesian decision strategies to maximize alignment accuracy.

Experimental results demonstrate that \MultiVLM substantially improves robustness against fabricated or misaligned table elements, achieving SOTA performance on OmniDocBench 1.5 and on the table subsets of CC‑OCR and PureDocBench. The integration of a custom 1B inference-efficient VLM further highlights the practicality of the proposed method for scalable deployment.

The study further underscores the value of combining multi‑perspective  content extraction with structured prior knowledge, offering a pathway toward more reliable document parsing systems. Beyond tables, the proposed paradigm might hold promise for other domains where outputs must satisfy strict logical or syntactic constraints.

\section{Limitations}

This work has several limitations.

First, employing multiple VLMs introduces additional computational overhead. Although the selected models are relatively lightweight and we introduce novel deployment strategies for shared encoders, the approach inherently trades efficiency for accuracy. Nevertheless, the combined three-model ensemble (totaling 3.1B parameters) achieves stronger benchmark performance than recent models of comparable or even larger scale.

Second,  experiments reveal that certain tables are incorrectly extracted by all three VLMs. In such case, the proposed method provides no mitigation, indicating a fundamental limitation of  model ensemble. In general, the capability of \MultiVLM is dependent on the capability of the VLMs chosen.

Third, while this study focuses on table, the proposed six-layer approach holds potential for extension to formula and diagram extraction. Exploring these directions is left for future work.

Finally, the proposed framework relies on a validation dataset to estimate variables in the Bayes formulation. If the distribution of this validation set diverges from  the benchmark, the resulting estimates may be inaccurate. Such misestimation can reduce, or in extreme cases negate the performance gains achievable through multi-VLM ensembling. 

\bibliography{custom}

\appendix

\section*{Appendix A}

The sets of table structure rules and table semantic rules we use are summarized in Table \ref{strucuture} and \ref{semantics}, respectively.
   \begin{table*}
    \centering
    \begin{tabular}{c | p{0.75\linewidth}}
    \toprule
    \textbf{No.} & \textbf{Table structure rules} \\
    \midrule
    1 & The output must contain valid HTML table control tags, e.g., <table>. \\
    2 & All \texttt{<td>}, \texttt{<th>}, and \texttt{<tr>} tags must be properly closed. \\
    3 & Column count must remain consistent across all rows, with column span attribute taken into account to ensure structural alignment. \\
    4 & Every column must contain the same number of rows, accounting for row span attribute where applicable. \\
    5 & HTML tags not related to tables should be filtered at the table structure level. \\
    6 & Rowspan and colspan attributes should not exceed the table’s dimensions, avoiding physically impossible layouts. \\
    \bottomrule
    \end{tabular}
        \caption{A set of table structure rules}
        \label{strucuture}
    \end{table*}

\begin{table*}
\centering
\begin{tabular} {c| p{0.23\linewidth} | p{0.65\linewidth}}
\toprule
\textbf{No.} & \textbf{Name} & \textbf{Table semantic rules} \\
\midrule
1 & Scale Constraints & The dimensions of the table must be bounded by a predefined threshold, e.g., 100 rows or 100 columns. Additionally, the values of rowspan and colspan attributes must not exceed this limit to prevent structural anomalies. \\
2& Content Repetition &  Cells containing repeated n-grams, specifically, sequences of 64 tokens or more, are considered degenerate and should be discarded. \\
3 & Boundary Validity &  Leading or trailing rows or columns that are entirely empty are semantically void and should be removed. \\

4 & Header Significance & A column consisting exclusively of numeric values must be anchored by a non-numeric header to provide necessary context.  \\
5 & Row Context & Rows primarily composed of numeric data must contain a categorical identifier (e.g., a label or key) in the first column. \\

6 & Formula Integrity & Cells containing structurally malformed mathematical formulas, e.g., unclosed LaTeX delimiters, are not preferred. \\
7 & Sparsity Threshold & Tables exhibiting an excessively low ratio of non-empty cells, for instance, below a preconfigured threshold, may indicate incomplete extraction or poor quality. \\

8 & Dimensionality Check & Tables with insufficient dimensions, e.g., fewer than 3 rows or 3 columns, often lack relational semantics. \\

9 & Alignment Consistency &  The spatial or logical boundaries of non-header cells must align with those of at least one corresponding header cell.  \\
10 & Ghost Columns & Columns where only the header contains content while all data cells are empty are deemed semantically meaningless. \\
	
11 & Type Homogeneity & Non-header cells within the same column must exhibit consistent data types, e.g., all numeric, all textual. \\
12 & Header Uniqueness & It is semantically implausible for multiple distinct numeric columns to share an identical header, if there is only one header row. \\
13 & Row Distinctness & Consecutive rows with identical cell values are redundant and likely artifacts of parsing errors. \\
14 & {Entity Consistency}& A column header indicating a specific entity type (e.g., "Year", "Country Code") must not contain values of conflicting types.\\
 \bottomrule
\end{tabular}
\caption{A set of computable table semantic rules}
\label{semantics}
\end{table*}

\newpage

\section*{Appendix B}

\begin{figure}[h]
    \centering
\includegraphics[width=0.8\textwidth]{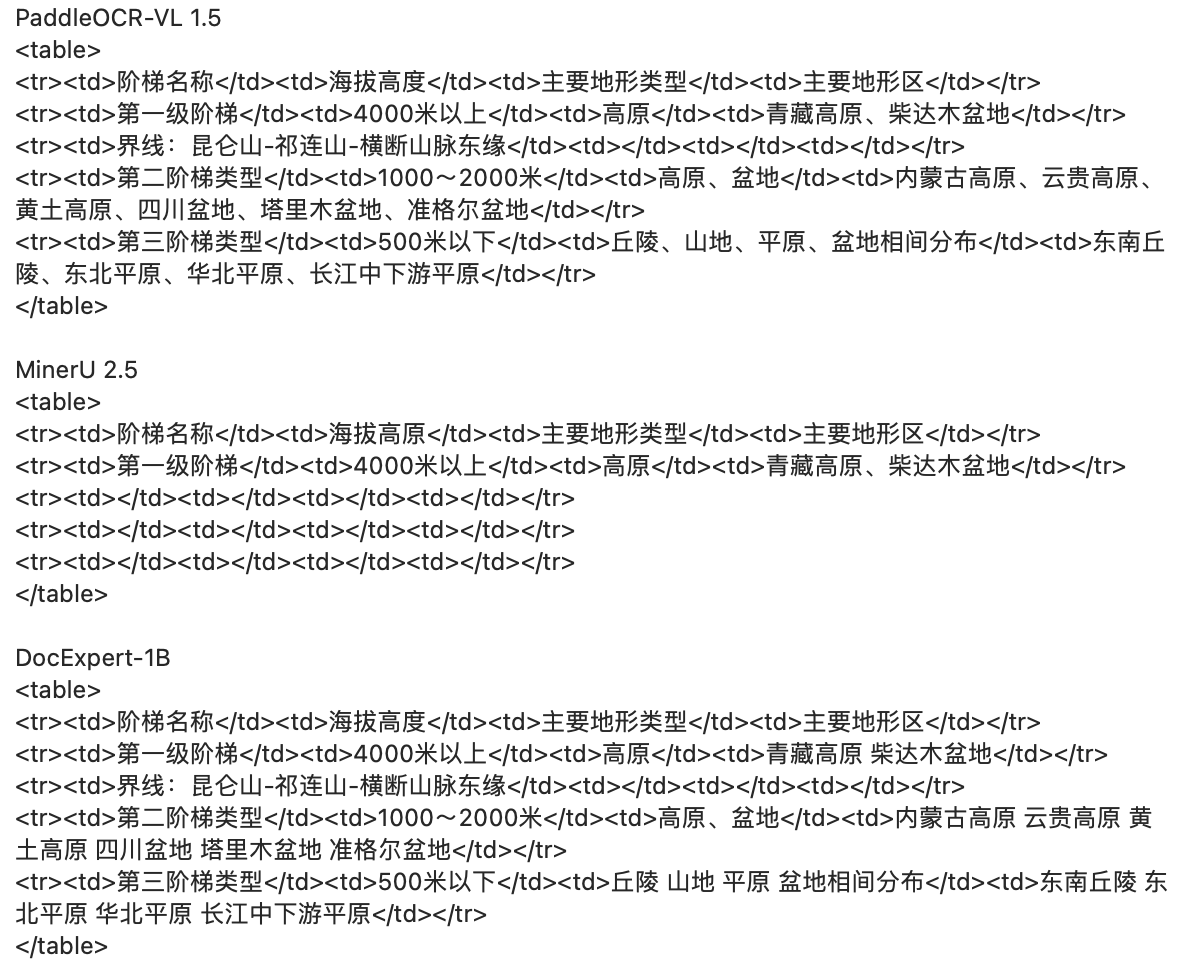} 
\caption{\MultiVLM component  result comparison}
\label{fig:compare}
\end{figure}

For the sample page in OmniDocBench, \texttt{notes \_1ba14cb325bc448f7201b20502ecf2b5\_60.jpg}, the extraction results for PaddleOCR-VL 1.5, MinerU 2.5, and DocExpert-1B, are shown in Fig. \ref{fig:compare}. For this case, no table structure rules are violated.  

For table semantic rules, 
MinerU 2.5 fails Rules 3, 7, and 13, while Paddle-VL 1.5 and DocExpert-1B do not. This results in MinerU 2.5  having a substantially lower posterior probability value compared to the other two methods. 

\vspace{48pt}

\end{document}